\documentclass[twoside,11pt]{article}

\usepackage[preprint]{jmlr2e}

\usepackage[utf8]{inputenc} 
\usepackage[T1]{fontenc}    
\usepackage{url}            
\usepackage{booktabs}       
\usepackage{amsfonts}       
\usepackage{xcolor}         
\usepackage[linesnumbered,ruled,vlined]{algorithm2e}
\usepackage{amsmath}
\usepackage{amssymb}
\usepackage{graphicx} 
\usepackage{authblk}

\usepackage{tikz}
\def\checkmark{\tikz\fill[scale=0.4](0,.35) -- (.25,0) -- (1,.7) -- (.25,.15) -- cycle;}

\ShortHeadings{Salesforce CausalAI Library}{Arpit et al.}

\title{Salesforce CausalAI Library: A Fast and Scalable Framework for Causal Analysis of Time Series and Tabular Data}

\author[1,*]{Devansh Arpit}
\author[1]{Matthew Fernandez}
\author[1,*]{Itai Feigenbaum}
\author[1]{Weiran Yao}
\author[1]{Chenghao Liu}
\author[1]{Wenzhuo Yang}
\author[1]{Paul Josel}
\author[1]{Shelby Heinecke}
\author[1]{Eric Hu}
\author[1]{Huan Wang}
\author[1]{Steven Hoi}
\author[1]{Caiming Xiong}
\author[2]{Kun Zhang}
\author[1,*]{Juan Carlos Niebles}
\affil[1]{AI Research, Salesforce}
\affil[2]{Carnegie Mellon University}
\affil[*]{Corresponding Authors: \texttt{devansharpit@gmail.com, jniebles@salesforce.com}}

\begin{document}
\maketitle

\begin{abstract}
We introduce the Salesforce CausalAI Library, an open-source library for causal analysis using observational data. It supports causal discovery and causal inference for tabular and time series data, of discrete, continuous and heterogeneous types. This library includes algorithms that handle linear and non-linear causal relationships between variables, and uses multi-processing for speed-up. We also include a data generator capable of generating synthetic data with specified structural equation model for the aforementioned data formats and types, that helps users control the ground-truth causal process while investigating various algorithms. Finally, we provide a user interface (UI) that allows users to perform causal analysis on data without coding. The goal of this library is to provide a fast and flexible solution for a variety of problems in the domain of causality. This technical report describes the Salesforce CausalAI API along with its capabilities, the implementations of the supported algorithms, and experiments demonstrating their performance and speed. Our library is available at \url{https://github.com/salesforce/causalai}.
\end{abstract}


\section{Introduction}
\label{sec_intro}
Causal inference aims at determining how a change in one part of a system affects another, in isolation, i.e., no other parts of the system are allowed to change independently. Such an inference is fundamentally different from predictions made by machine learning models, which are based on correlation between variables. The reason behind this difference is that correlation does not necessarily imply causation. As a simple example, consider two discrete random variables, $X$ and $Y$. $X$ can independently take two states-- $-1$ and $+1$ with probability $0.5$ each. $Y$ on the other hand takes the state $0$ when $X$ is $-1$, and states $-1$ and $+1$ with probability $0.5$ each, when $X$ is $+1$. By design, $X$ causes $Y$. However, the correlation between the two variables is $0$. Another example on the other end of the spectrum is a scenario in which a third variable $Z$ is a common cause of $X$ and $Y$, and there is no causal link between the latter two. In this case, it is likely that $X$ and $Y$ are correlated. However, by design, any isolated change in $X$ cannot cause changes in $Y$, since $Z$ is fixed. 

The above examples illustrate the fundamental limitations of using correlation based models for the problem of predicting the causal impact of one variable on another. This problem has important applications in several domains such as sales, medicine, diagnosis, etc. For instance, a business might be interested in finding out whether more customers would be likely to purchase their products if they offered a discount, versus if they showed ads on the television. As another example, finding out if certain chemicals cause harmful effects on health can benefit the broader society. This type of knowledge facilitates interventions, which are actionable items that can be used to change future outcomes, as opposed to correlation based machine learning models, which are typically used for automation. Causal analysis tools help us discover which variables in a system can be intervened to achieve the desired outcome for a particular variable of interest.


Given the importance of this problem, several libraries exist on causal discovery and causal inference \citep{kalainathan2019causal,sharma2020dowhy,Beaumont_CausalNex_2021}. However, certain limitations still exist in existing libraries, such as computationally heavy implementation, and the lack of support for different data types and code-free user interface, which are required for a unified end-to-end system that supports a fast and easy to use system for the various types of problems in the domain of causal analysis. 

\begin{figure}[t]
    \centering
    \includegraphics[width=1\linewidth, trim=0.2in 2.1in 0.2in 1.in, clip]{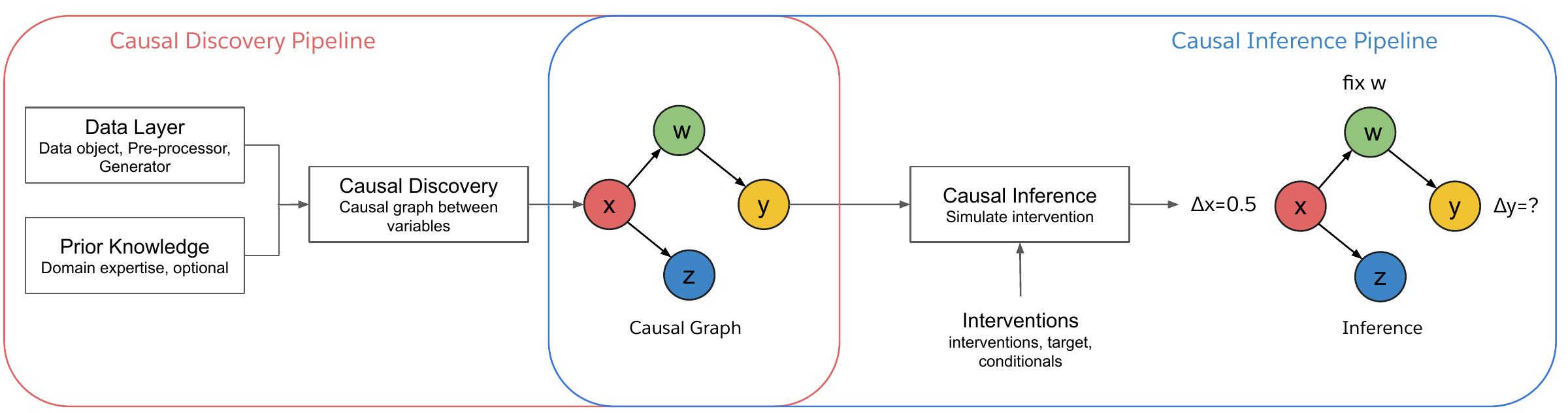}
    \caption{Salesforce CausalAI Library Pipeline. We support causal discovery and causal inference. The causal discovery module takes as input a data object (containing observational data) and a prior knowledge object (containing any expert partial prior knowledge, optional), and outputs a causal graph. The causal inference module takes a causal graph as input (which could be directly provided by the user or estimated using the causal discovery module) along with the user specified interventions, and outputs the estimated effect on the specified target variable. \label{fig:pipeline}}
\end{figure}

We introduce the Salesforce CausalAI Library, a Python library for causal analysis that supports causal discovery and causal inference using observation data. The Salesforce CausalAI library pipeline is shown in figure \ref{fig:pipeline}. Some of the key features of our library are:
\begin{itemize}
    \item \textbf{Algorithms}: Support for causal discovery, causal inference, Markov blanket discovery, and Root Cause Analysis (RCA).
    \item \textbf{Data}: Causal analysis on tabular and time series data, of discrete, continuous, and heterogeneous types.
    \item \textbf{Missing Values}: Support for handling missing/NaN values in data.
    \item \textbf{Data Generator}: A synthetic data generator that uses a specified structural equation model (SEM) for generating tabular and time series data. This can be used for evaluating and comparing different causal discovery algorithms since the ground truth values are known.
    \item \textbf{Distributed Computing}: Use of multi-processing using the Ray \citep{moritz2018ray} library, that can be optionally turned on by the user when dealing with large datasets or number of variables for faster compute.
    \item \textbf{Targeted Causal Discovery}: In certain cases, we support targeted causal discovery, in which the user is only interested in discovering the causal parents of a specific variable of interest instead of the entire causal graph. This option reduces computational overhead.
    \item \textbf{Visualization}: Visualize tabular and time series causal graphs.
    \item \textbf{Domain Knowledge}: Incorporate any user provided partial prior knowledge about the causal graph in the causal discovery process.
    \item \textbf{Benchmarking}: Benchmarking module for comparing different causal discovery algorithms, as well as evaluating the performance of a particular algorithm across datasets with different challenges.
    \item \textbf{Code-free UI}: Provide a code-free user interface in which users may directly upload their data and perform their desired choice of causal analysis algorithm at the click of a button.
\end{itemize}




\section{Related Work}

Table \ref{tab:related_works} summarizes the main features supported by the Salesforce CausalAI library versus the existing libraries for causal analysis. The key differentiating features of our library are parallelization and a user interface (UI), that are aimed at making causal analysis more scalable and user friendly. Our UI allows the users to run causal discovery and causal inference algorithms on their local machines without the need to write code. In terms of parallelization, we use the Python Ray library \citep{moritz2018ray} (optional) such that the implementation of the algorithm is tied to it, which makes the user's interaction with the API simple, i.e., the user can simply specify as an argument whether or not to use multi-processing. Tigramite also supports parallelization, but uses MPI4Py \citep{dalcin2021mpi4py}, which we found to be slower compared to Ray. Additionally, in Tigramite's implementation of the PC algorithm, the process of finding the causal neighbors of a variable is run in parallel for each variable. Note that each causal neighbor discovery may require an exponential number (in terms of the number of variables) of conditional independence (CI) tests. Our implementation on the other hand runs the CI tests themselves within each causal neighbor discovery process in parallel. This makes our implementation more efficient. CausalNex on the other hand uses Pytorch for parallelization.

\begin{table}[]
\centering
\small
\begin{tabular}{l|c|c|c|c|c|c|c|c|c|c}
Library                        & CI & DK & Time Series & Tabular & Parallel & UI & MBD & RCA & Benchmark \\ \hline
Tetrad                        & \checkmark                & \checkmark                & \checkmark           & \checkmark                  &     -     & \checkmark     & - & - & -              \\
Causal-learn                  &         -         & \checkmark                & \checkmark           & \checkmark               &  -        & -              & - & - & -        \\
Tigramite                     & \checkmark                &          -        & \checkmark           &      -         &    \checkmark      & \checkmark              & - & - & -        \\
CausalNex                      & \checkmark                & \checkmark                &       -      & \checkmark              & \checkmark        &           -      & - & - & -        \\
DoWhy                          & \checkmark                & \checkmark                &        -     & \checkmark             & -       &        -          & - & - & -       \\
CausalAI (ours)           & \checkmark                & \checkmark                & \checkmark           & \checkmark               & \checkmark        & \checkmark     & \checkmark & \checkmark &        \checkmark
\end{tabular}
\caption{\small A comparison of features supported by Salesforce CausalAI library vs existing libraries for causal analysis. Note that all libraries support causal discovery. \textbf{CI}: Causal Inference, \textbf{DK}: Domain Knowledge, \textbf{UI}: User Interface,  \textbf{MBD}: Markov Blanket Discovery, and \textbf{RCA}: Root Causal Analysis.\label{tab:related_works}}
\end{table}

\section{Library Architecture and API Description}

The Salesforce CausalAI Library API is composed of four main components-- data layer, prior knowledge, causal discovery, and causal inference, as described in subsections below. The data layer includes data generator, data transform and data specification modules. The prior knowledge layer helps the user specify any partial knowledge about the causal relationship between variables. The causal discovery layer aims at retrieving the underlying causal graph from observational data. Finally, the causal inference layer aims at estimating the average treatment effect (ATE) and conditional ATE (CATE) of interventions on a subset of variables on a specified target variable. Aside from these components, we also support causal graph visualization and evaluation of the correctness of the estimated causal graph when the ground truth graph is available.

\subsection{Data Layer}

\textbf{Data Generator}

The function \texttt{DataGenerator} can be used to generate synthetic time series and tabular data according to a user specified additive noise structural equation model (SEM). As an example, consider a tabular data system with random variables $A$, $B$ and $C$, then an instance of an additive noise SEM is,
\begin{align}
    A &= \mathcal{N}_1()\\
    B &= k_1 \times F_1(A) + \mathcal{N}_2()\\
    C &= k_2 \times F_2(A) + k_3 \times F_3(B) + \mathcal{N}_3()
\end{align}
where $\mathcal{N}_i$'s and $F_i$'s are callable functions, $k_i$'s are constants, all of which are user specified. This SEM is passed as the following dictionary $\texttt{sem}=\{A:[], B: [(A, k_1, F_1)], C: [(A, k_2, F_2), (B, k_3, F_3)]\}$. The noise functions $\texttt{noise\_fn}=[\mathcal{N}_1, \mathcal{N}_2, \mathcal{N}_3]$ is passed as a list. The procedure for generating data from this SEM is to traverse a topologically sorted order of variables in the corresponding causal graph and sample data-- first sample the values of $A$, since it has no parents, then $B$ and then $C$. The number of data samples is specified by \texttt{T} (int). The user may provide a random seed value \texttt{seed} (int), for reproducibility. \texttt{DataGenerator} accepts \texttt{intervention} if the goal is to generate data from a specified SEM while intervening on certain variables. This is passed as a dictionary, where keys are the variable names to be intervened, and values are 1D NumPy arrays of length equal to \texttt{T}. The way this is achieved is identical to the previous case, except that \texttt{DataGenerator} intervenes on the intervention variables whenever they are encountered during the graph traversal. Finally, the user may specify whether the generated data should be discrete or continuous via \texttt{discrete} (bool). In the discrete case, the \texttt{DataGenerator} function first generates continuous data, and then discretizes them by binning (the number of states can be specified via \texttt{nstates}). 

There are cases where it is desirable to generate data using a given SEM, both with and without interventions. To achieve this, the \texttt{DataGenerator} function is called twice, once without the intervention argument, and another with the intervention argument. This gives the user access to observational data from that SEM, but also provides the ground truth value of variables after a subset of variables are intervened. This is useful, for instance, when we want to evaluate the accuracy of causal inference algorithms. In such cases, it is important to specify the same random seed to the \texttt{DataGenerator} function in both the calls.
\\\\
\textbf{SEM Generator}
For each, tabular and time series data, we support two modules (total four) for generating random causal graphs along with the corresponding SEM. 

For time series data, we support \texttt{GenerateRandomTimeseriesSEM} and \texttt{GenerateSparseTimeSeriesSEM}. Both modules take as input \texttt{var\_names}, \texttt{max\_lag}, \texttt{seed}, \texttt{fn}, and \texttt{coef}, which respectively specify the list of variable names (strings), maximum time lag (positive integer) between a parent and child node in the causal graph, a random seed for reproducibility, the non-linearity function (similar to $F_1$ in the example SEM equation above), and the coefficient (similar to $k_1$ in the example SEM equation above) multiplied to the parent variable value (after the non-linearity). In addition, \texttt{GenerateRandomTimeseriesSEM} takes the input \texttt{max\_num\_parents}, which specifies the maximum number of parents that each node can have, and \texttt{GenerateSparseTimeSeriesSEM} takes the input \texttt{graph\_density} which specifies the probability with which an edge is sampled between any two nodes. The main difference between the two modules is the argument \texttt{max\_num\_parents} vs \texttt{graph\_density}, which differentiates in the way the random causal graph is generated.

For tabular data, we similarly support \texttt{GenerateRandomTabularSEM} and \texttt{GenerateSparseTabularSEM}. The arguments used by both the methods are similar to their time series counterpart, except that they do not take \texttt{max\_lag} as input.
\\\\
\textbf{Data Transform}

For time series data, three transform modules are supported-- \texttt{StandardizeTransform}, \texttt{DifferenceTransform}, and \texttt{Heterogeneous2DiscreteTransform}. \texttt{StandardizeTransform} subtracts and divides each element with the corresponding feature-wise mean and standard deviation respectively. \texttt{DifferenceTransform} performs temporal differencing, i.e., computes the difference between two time-steps separated by a given interval. This interval can be specified by the user. Note that in case there are NaN values in the data array, the output of these transformation only contains NaNs at the location of the original NaNs. These modules use NumPy array format as input and output. Finally, multiple arrays may be passed to these modules in the case of disjoint time series data (E.g. one time series is a daily record from January to March 2001, and another from January to March 2002). All the modules have a \texttt{fit} method that takes the data arrays as input, which are used for computing the transformations. All modules have a \texttt{transform} method, which takes as input the data arrays to be transformed using the said transformation.

For tabular data, \texttt{StandardizeTransform} and \texttt{Heterogeneous2DiscreteTransform} are supported, which works identically to the time series case.

For both tabular and time series data, we support heterogeneous variables (mixed continuous and discrete variables) for causal discovery. To achieve this, we support a data transform module called \texttt{Heterogeneous2DiscreteTransform} for both tabular and time series data, which converts a mixed variables data into an all discrete variable data. The user needs to specify \texttt{nstates} (int) during module initialization to specify the number of states that is used during the discretization process. Additionally, the \texttt{fit} method takes as input \texttt{discrete}, which is a Python dictionary with variable names as keys, and True/False as values, specifying whether the corresponding variable is discrete or continuous.
\\\\
\textbf{Data Object}

In order to feed observational data to the causal discovery algorithms in our API, the raw data-- NumPy arrays and a list of variable names (optional), is used to instantiate a CausalAI data object. Note that any data transformation must be applied to the NumPy array prior to instantiating a data object. For time series and tabular data, \texttt{TimeSeriesData} and \texttt{TabularData} must be initialized with the aforementioned data respectively. For time series data, multiple arrays may be passed to the \texttt{TimeSeriesData} class in the case of disjoint time series data (E.g. one time series is a daily record from January to March 2001, and another from January to March 2002). Users may also optionally specify if the data contains or may contain NaNs, so they may be handled. These data objects contain some useful methods (E.g. for extracting the data corresponding to the parents of a variable for a given observation or time step), but they are mainly designed to be used internally by the causal discovery algorithms, and users may treat the data object as a blackbox.

\subsection{Prior Knowledge}

The module \texttt{PriorKnowledge} can be optionally used to specify any partial prior knowledge about the causal graph. This module is used in causal discovery algorithms supported by our API. Specifically, there are six types of prior knowledge that users may specify in the form of arguments to the \texttt{PriorKnowledge} class-- \texttt{forbidden\_links}, \texttt{existing\_links}, \texttt{root\_variables}, \texttt{leaf\_variables}, \texttt{forbidden\_co\_parents} and \texttt{existing\_co\_parents}. The argument \texttt{forbidden\_links} should specify which variables cannot be the parents of a variable. The argument \texttt{existing\_links} should specify which variables are known to be the parents of a variable. The argument \texttt{root\_variables} should specify which variables cannot have any causal parents. The argument \texttt{leaf\_variables} should specify which variables cannot have any causal children. Finally, \texttt{forbidden\_co\_parents} and \texttt{existing\_co\_parents} are used for Markov Blanket Discovery: they follow the same format as \texttt{forbidden\_links} and \texttt{existing\_links}, but specify which variables cannot/must be co-parents of a variable. The module performs consistency checks to prevent self-contradictions in the provided prior knowledge.

Note that these specifications are accepted in the same format for both tabular and time series data. Specifically, in the time series case, the user may specify that $A \rightarrow B$ is forbidden, which implies that no time lagged or instantaneous state of variable $A$ can cause $B$ at any time step. Further, \texttt{existing\_links} are not utilized for time series data since time lag information cannot be specified in \texttt{PriorKnowledge}. The reasoning behind this design choice is that it is more likely that domain experts may know whether two time series variables are causally related, but less likely that they know which specific time lag of one variable causes another.

\texttt{PriorKnowledge} is also capable of expanding on the information given to it. If the argument \texttt{fix\_co\_parents} is set to True, \texttt{PriorKnowledge} attempts to supplement the given information regarding co-parent relationships by making deductions based on its other inputs. First, \texttt{existing\_co\_parents} is updated with co-parent relationships implied by \texttt{existing\_links}. Next, \texttt{forbidden\_co\_parents} is updated with co-parent relationship forbidden by \texttt{leaf\_variables}. Specifically, leaf variables are added as forbidden co-parents for all variables for which the second type of update is considered, specified by the \texttt{var\_names} argument. Finally, since co-parenting is a symmetric relationship, both \texttt{forbidden\_co\_parents} and \texttt{existing\_co\_parents} are expanded to be symmetric: this last update occurs regardless of the value of \texttt{fix\_co\_parents}.

\subsection{Causal discovery}
\label{sec_cd_api}

All the supported causal discovery algorithm classes take a data object (\texttt{TimeSeriesData} or \texttt{TabularData}) and optionally \texttt{PriorKnowledge} as input during instantiation. In order to make a decision on whether a causal edge exists between two variables or not, all implemented algorithms internally make use of statistical hypothesis testing, specifically by computing p-values. Further, all these classes have a \texttt{run} method that takes \texttt{pvalue\_thres} as input argument (among other case specific inputs), and returns a causal graph (potentially with some undirected edges) along with the strength of the discovered causal edges and their p-values.

\subsubsection{Time Series Data}
For time series data, the \texttt{run} method takes \texttt{max\_lag} argument, a non-negative integer, which specifies the maximum possible time lag allowed for causal parents.

\textbf{Continuous data}: For time series continuous data, we currently support PC algorithm \citep{spirtes2000causation}, Granger causality \citep{granger1969investigating}, and VARLINGAM \citep{hyvarinen2010estimation}. Granger causality, and VARLINGAM only support linear causal relationship between variables, while PC in general supports non-linear relationships. For targeted causal discovery (see section \ref{sec_intro}), \texttt{PCSingle} and \texttt{GrangerSingle} retrieve the causal parents of a given target variable. Their \texttt{run} method takes as additional argument \texttt{target\_var}, which specifies the target variable name. For full causal discovery, \texttt{PC}, \texttt{Granger} and \texttt{VARLINGAM} classes should be used. Note that of the three algorithms, only \texttt{VARLINGAM} supports instantaneous causal edge discovery in our library. 

Finally, \texttt{PCSingle} and \texttt{PC} API take two additional arguments-- \texttt{max\_condition\_set\_size} and \texttt{CI\_test}. Since the PC algorithm performs conditional independence (CI) tests to find causal dependence between two variables, we need to specify which test to use. This is done through the \texttt{CI\_test} argument. Currently we support two choices-- \texttt{PartialCorrelation} and \texttt{KCI}, but users may also specify their own CI test class. \texttt{PartialCorrelation} assumes linear causal relationship between variables while \texttt{KCI} allows non-linear relationships, depending on the kernel used (see tutorials in our repository for examples). The argument \texttt{max\_condition\_set\_size} on the other hand specifies the maximum size of condition set to be used during CI tests (the upper limit of \texttt{max\_condition\_set\_size} is $N-2$, where $N$ is the total number of variables). Larger values make the PC algorithm exponentially slower, but typically more accurate. We also support specifying \texttt{max\_condition\_set\_size} as \texttt{None}. In this case, CI tests are performed using all $N-2$ variables\footnote{Note that performing exactly one CI test for a pair of variables using a condition set containing all the remaining $N-2$ variables is not fundamentally incorrect in the case when only time lagged parents are considered. More discussion on this can be found in \ref{sec_pc_implem}}. While this can speed up the PC algorithm in the worst case, the results may be less accurate. Ideally, we recommend using a small integer (default value is 4) as a trade-off between speed and accuracy.

\textbf{Discrete data}: For time series discrete data, \texttt{PCSingle} and \texttt{PC} can be used. In this case, we support the \texttt{DiscreteCI\_tests} class, in which one of the following CI tests can be specified in the \texttt{method} argument-- Chi-squared test (\texttt{pearson}, default), log-likelihood test (\texttt{log-likelihood}), Modified Log-likelihood (\texttt{mod-log-likelihood}), Freeman-Tukey Statistic (\texttt{freeman-tukey}), and Neyman's statistic (\texttt{neyman}). The additional arguments to be specified to the \texttt{run} method for these API's are similar to the continuous case above.

\subsubsection{Tabular Data}

\textbf{Continuous Data}: For tabular continuous data causal discovery, we support PC, GES, LINGAM, and GIN algorithms, as well as Markov Blanket discovery algorithms (described in the next paragraph). All algorithms except PC support only linear causal relationship between variables, while PC in general supports non-linear relationships. GIN on the other hand supports causal discovery in data with hidden confounding variables. All the methods support full causal discovery and only PC supports targeted causal discovery. The \texttt{run} method in all these algorithms' API uses the \texttt{pvalue\_thres} input argument, except the GES algorithm. Finally, the \texttt{run} method of GES takes the following input arguments: \texttt{A0}, and \texttt{phases}. \texttt{phases} specifies which of the GES phases are run. Supported values are \texttt{forward}, \texttt{backward}, and \texttt{turning}. By default, all are used. \texttt{A0} on the other hand specifies the initial state of the causal graph used by GES. \texttt{A0}[i,j] $!= 0$ implies $i \rightarrow j$ and (\texttt{A0}[i,j] $!= 0$  $\&$ \texttt{A0}[j,i] $!= 0$) implies $i - j$.

We also support Markov Blanket Discovery, specifically, the Grow-Shrink algorithm. The \texttt{GrowShrink} class takes the same arguments as the causal discovery algorithm classes, as well as an \texttt{update\_shrink} boolean variable indicating whether or not to use the most current Markov Blanket estimate for CI testing during the second half of the algorithm. The \texttt{run} method estimates the Markov Blanket of the \texttt{target\_var} input argument, using the \texttt{pvalue\_thres} input argument for conditional independence testing.

\textbf{Discrete Data}: For tabular discrete data causal discovery, we support PC. The API is similar to the time series API.

\subsection{Causal inference}

We support the class \texttt{CausalInference} to perform causal inference for tabular and time series data, which can be used to compute average treatment effect (ATE) and conditional ATE (CATE). This class can be initialized using a 2D NumPy data array \texttt{data}, a list of variable names \texttt{var\_names}, a Python dictionary specifying the causal graph \texttt{causal\_graph} corresponding to the data, a \texttt{prediction\_model} to be used for learning the mapping function between different variables, the argument \texttt{discrete} (bool) specifying whether the data is continuous or discrete. Another argument \texttt{use\_multiprocessing} (bool) may also be specified at initialization which can be used to speed up computation. Typically, multi-processing has an advantage when \texttt{prediction\_model} is a non-linear model.

To perform ATE upon initialization, the method \texttt{ate} should be called with the arguments \texttt{target\_var} and \texttt{treatments}. Here \texttt{target\_var} is the name of the variable on which ATE needs to be estimated, and \texttt{treatments} is a list of Python dictionary, or a dictionary specifying the intervention variables and their corresponding treatment and control values. Specifically, this Python dictionary has three keys called \texttt{var\_name}, \texttt{treatment\_value}, \texttt{control\_value}.

To perform CATE upon initialization, the method \texttt{cate} should be called. This method takes two additional arguments in addition to those described for the \texttt{ate} method above. They are \texttt{conditions} and \texttt{condition\_prediction\_model}.

\subsection{Applications: Root Cause Analysis}

We support the class \texttt{RootCauseDetector} and the class \texttt{TabularDistributionShiftDetector} to perform root cause analysis for  anomalies in time-series and tabular data using the PC algorithm. Both classes can be initialized using pre-processed TabularData objects, a list of variable names \texttt{var\_names}, a variable indicating the time-varying context \texttt{time\_metric\_name} in \texttt{var\_names} for \texttt{RootCauseDetector}, or a variable indicating the domain index~, i.e., \texttt{domain\_index\_name} for \texttt{TabularDistributionShiftDetector}, and optional prior knowledge of the causal graph \texttt{prior\_knowledge}. 

To perform root cause analysis, the method \texttt{run} should be called with the arguments \texttt{pvalue\_thres}, which is a float specifying the significance level for the conditional independence test, \texttt{max\_condition\_set\_size}, which is an integer specifying the maximum number of parent nodes the PC algorithm considers for the causal search, and \texttt{return\_graph}, which is a boolean variable determining whether or not to return the discovered causal graph together with the identified root causes. The method \texttt{run} returns \texttt{root\_causes}, which is a Python set, and the \texttt{graph}, which is a Python dictionary with keys being the parents and values being the children of the parents in the recovered causal graph. 

\subsection{Benchmarking}

The benchmarking module supports methods that evaluate causal discovery algorithms for tabular and time series data against various challenges-- sample complexity, variable complexity, noise function. signal-to-noise ratio, and graph density. For the case of time series, we additionally support evaluation against maximum time lag between causally connected nodes. Users can use either synthetically generated data, or provide their own data for benchmarking. 

The default evaluation metrics supported by this module are Precision, Recall, F1 Score, and Time Taken by the algorithm. There is also an option for users to 
include their own custom metrics when calling the benchmarking module.

We provide support for a default set of causal discovery algorithms. Users also have the option to include their own algorithm when calling the 
benchmarking module.

There are two options in terms of data-- either use synthetically generated data, or provide custom data. 

The supported benchmarking modules for time series are \texttt{BenchmarkContinuousTimeSeries} and
\texttt{BenchmarkDiscreteTimeSeries} for continuous and discrete data respectively.

The supported benchmarking modules for tabular data are
\texttt{BenchmarkContinuousTabular} and
\texttt{BenchmarkDiscreteTabular} for continuous and discrete data respectively.

The discrete modules for both tabular and time series data support the methods \\
\texttt{benchmark\_variable\_complexity},  \texttt{benchmark\_sample\_complexity}, and \\
\texttt{benchmark\_graph\_density}. The continuous modules additionally support the methods \texttt{benchmark\_noise\_type}, and
\texttt{benchmark\_snr}.

Finally, the time series modules additionally support the method \texttt{benchmark\_data\_max\_lag}.

\section{Causal Discovery Algorithms}

Causal discovery aims at finding the underlying directed causal graph from observational data, where the variables (or features) are treated as nodes in the graph, and the edges are unknown. For two variables $A$ and $B$ in a graph, the edge $A \rightarrow B$ denotes $A$ causes $B$. Observational data is simply a set of observations recorded in the past without actively making any interventions. Typically, finding causal relationships between variables would require performing interventions. But under certain assumptions, it is possible to extract the underlying causal relationships between variables from observational data as well.
In this section, we describe some of such algorithms that are supported by our library, their assumptions, and their implementation details.

\subsection{PC Algorithm}
\label{sec_pc_implem}

The Peter-Clark (PC) algorithm \citep{spirtes2000causation} is one of the most general purpose algorithms for causal discovery that can be used for both tabular and time series data, of both continuous and discrete types. Briefly, the PC algorithm works in two steps, it first identifies the undirected causal graph, and then (partially) directs the edges. In the first step, we check for the existence of a causal connection between every pair of variables by checking if there exists a condition set (a subset of variables excluding the two said variables), conditioned on which, the two variables are independent. In the second step, the edges are directed by identifying colliders. Note that the edge orientation strategy of the PC algorithm may result in partially directed graphs. In the case of time series data, the additional information about the time steps associated with each variable can also be used to direct the edges.

The PC algorithm makes four core assumptions: (1) Causal Markov condition, which implies that two variables that are d-separated in a causal graph are probabilistically independent, (2) faithfulness: no conditional independence can hold unless the Causal Markov condition is met, (3) no hidden confounders, and (4) no cycles in the causal graph. For time series data, it makes the additional assumption of stationarity: the properties of a random variable are agnostic to the time step.

Our implementation of the PC algorithm for time series supports lagged causal relationship discovery. Consider a dataset with $N$ variables. To decide whether there is a causal edge between two variables, typically, this algorithm involves searching for condition sets (this set always excludes the said two variables), for which the two variables are conditionally independent. If a set is found at any point, the search is terminated for the said two variables, and it is concluded that there is no causal edge between the them, otherwise there is one. As can be seen, in the worst case, this search may require an exponential number of steps to determine the existence of a causal edge between each pair of variables. In our implementation, we support the following ways to speed up this algorithm:

\noindent\textbf{1.} The user may specify \texttt{max\_condition\_set\_size} (see section \ref{sec_cd_api}) to be a small integer (default is 4) to terminate the search once the condition set set size \texttt{max\_condition\_set\_size} is reached. The intuition behind this is idea is that one may expect the causal graph to be sparsely connected, in which case, \texttt{max\_condition\_set\_size} can be the maximum number of edges attached to a node.

\noindent\textbf{2.} Multi-processing: For any given condition set size, all the conditional independence tests are independent of one another. We allow the user to exploit this fact and run these tests in parallel. This is optional, and is beneficial when the number of variables or samples are large. Since instantiating and terminating a Ray object (we use the Ray library for multi-processing) adds a small overhead (typically a few seconds), using multi-processing may be slower for small datasets.

\noindent\textbf{3.} In the case of time series specifically, we also support a mode in which conditional independence tests are performed using all the $N-2$ variables. This can be done by setting $\texttt{max\_condition\_set\_size}=\texttt{None}$. This way, the number of CI tests reduces from exponential to one, per pair of variables, thus saving compute time significantly in the worst case. Note that this technically makes sense in our implementation of PC because it supports only time-lagged causal discovery (assuming no contemporaneous relationships). Therefore, no colliders are possible for the two variables under test, since one of the variable is always the current time step, and all other variables are past time steps. However, note that using this full CI test may reduce the power of the independence test and result in less accurate causal discovery, as described in \citep{runge2019detecting}. To avoid this, one solution is to set \texttt{max\_condition\_set\_size} to be a small integer (similar to the description above). The difference in the time series case is that once the condition sets of size \texttt{max\_condition\_set\_size} is reached, our implementation automatically performs the full CI test using all the variables that do not get eliminated as candidate parents during the greedy search process. Thus our implementation serves as a middle ground between the traditional PC algorithm implementation and the implementation with full CI test.

\subsection{Granger Causality}
\label{sec_granger}

Granger causality \citep{granger1969investigating} can be used for causal discovery in time series data without contemporaneous causal connections. The intuition behind Granger causality is that for two time series random variables $X$ and $Y$, if including the past values of $X$ to predict $Y$ improves the prediction performance, over using only the past values of $Y$, then $X$ causes $Y$. In practice, to find the causal parents of a variable, this algorithm involves performing linear regression to predict that variable using the remaining variables, and using the regression coefficients to determine the causality.

Granger causality assumes: (1) linear relationship between variables, (2) covariance stationary: a temporal sequence of random variables all have the same mean and the covariance between the random variables at any two time steps depends only on their relative positions, and (3) no hidden confounders.

This algorithm supports lagged causal relationship discovery. Since this algorithm involves running regression problems independently for each variable for the full causal graph, our implementation optionally allows the user to perform these optimization tasks using multi-processing for large datasets to speed up causal discovery.

\subsection{VARLINGAM}

VARLINGAM \citep{hyvarinen2010estimation} can be used for causal discovery in time series data with contemporaneous causal connections. This algorithm can be broadly divided into two steps. First, we estimate the time lagged causal effects using vector autoregression. Second, we estimate the instantaneous causal effects by applying the LiNGAM \citep{shimizu2006linear} algorithm on the residuals of the previous step, where LiNGAM exploits the non-Gaussianity of the residuals to estimate the instantaneous variables' causal order.

This algorithm makes the following assumptions: (1) linear relationship between variables, (2) non-Gaussianity of the error (regression residuals), (3) no cycles among contemporaneous causal relations, and (4) no hidden confounders. We do not support multi-processing for this algorithm.

\subsection{Greedy Equivalence Search (GES)}
This is an algorithm for tabular data causal discovery. Greedy Equivalence Search (GES, \citet{chickering2002optimal}) heuristically searches the space of causal Bayesian network and returns the model with highest Bayesian score it finds. Specifically, GES starts its search with the empty graph. It then performs a forward search in which edges are added between nodes in order to increase the Bayesian score. This process is repeated until no single edge addition increases the score. Finally, it performs a backward search that removes edges until no single edge removal can increase the score.

This algorithm makes the following assumptions: 1. observational samples are i.i.d.; 2. linear relationship between variables with Gaussian noise terms; 3. Causal Markov condition, which implies that two variables that are d-separated in a causal graph are  probabilistically independent; 4. faithfulness, i.e., no conditional independence can hold unless the Causal Markov condition is met; 5. no hidden confounders.

\subsection{LINGAM}
LINGAM \citep{shimizu2006linear} can be used for causal discovery in tabular data. The algorithm works by first performing independent component analysis (ICA) on the observational data matrix $X$ (variables $\times$ samples) to extract the mixing matrix $A$ over the independent components (noise matrix) $E$ (same size as $X$), i.e. solving $X=AE$. Then their algorithm uses the insight that to find the causal order, each sample $x$ can be decomposed as, $x = Bx + e$, where $B$ is a lower triangular matrix and $e$ are the independent noise samples. Noticing that $B = (I - A^{-1})$, we solve for $B$, and find the permutation matrix $P$, such that $PBP^\prime$ is as close to a lower triangular matrix as possible.

This algorithm makes the following assumptions: 1. linear relationship between variables, 2. non-Gaussianity of the error (regression residuals), 
3. The causal graph is a DAG, 4. no hidden confounders. We do not support multi-processing for this algorithm.

\subsection{Generalized Independence Noise (GIN)}
Generalized Independent Noise (GIN, \citet{xie2020generalized}) is a method for causal discovery for tabular data when there are hidden confounder variables.

Let X denote the set of all the observed variables and L the set of unknown ground truth hidden variables. 
Then this algorithm makes the following assumptions:
1. There is no observed variable in X, that is an ancestor of any latent variables in L.
2. The noise terms are non-Gaussian.
3. Each latent variable set L' in L, in which every latent variable directly causes the same set of 
observed variables, has at least 2Dim(L') pure measurement variables as children.
4. There is no direct edge between observed variables.

\subsection{Markov Blanket Discovery: Grow-Shrink}

The Grow-Shrink algorithm \citep{margaritis1999bayesian} can be used for discovering the minimal Markov blanket (MB) of a target variable in tabular data. A MB is a minimal conditioning set making the target variable independent of all other variables; under the assumption of faithfulness, which we make here, the MB is unique and corresponds to the set of parents, children and co-parents of the target variable. The MB can be used for feature selection.

The Grow-Shrink algorithm operates in two phases, called growth and shrink. The growth phase first adds to the MB estimation variables unconditionally dependent on the target variable, then conditions on those variables and adds the conditionally dependent variables to the estimation. Assuming perfect conditional independence testing, this yields a superset of the actual MB. The shrink phase then removes from the estimated MB variables independent from  he target variable conditional on all other variables in the MB estimation. The algorithm does not partition the estimated MB into parents/children/co-parents.

The assumptions we make for the growth-shrink algorithm are: 1. Causal Markov condition, which implies that two variables that are d-separated in a causal graph are probabilistically independent, 2. faithfulness, i.e., no conditional independence can hold unless the Causal Markov condition is met, 3. no hidden confounders, and 4. no cycles in the causal graph.

\section{Causal Inference Algorithms}


Causal inference involves finding the numerical estimate of intervening one set of variables, on another variable. This type of inference is fundamentally different from what machine learning models do when they predict one variable given another as input, which is based on correlation between the two variables found in the training data. In contrast, causal inference tries to estimate how a change in one variable propagates to the target variable while traversing the causal graph from the intervened variable to the target variable along the directed edges. This means that even if two or more variables are correlated, intervening one may not have any effect on another variable if there is no causal path between them.

Specifically, in this library, we support estimating average treatment effect (ATE), conditional ATE (CATE) and Counterfactuals. Suppose we are interested in estimating the ATE of some intervention of a set of variables denoted as $X$, on a variable $Y$. The treatment and control values of $X$ are denoted as $x_t$ and $x_c$. Then ATE is defined as,
\begin{align}
    \label{eq_ate}
    \texttt{ATE} = \mathbb{E}[Y | \texttt{do}(X=x_t)] - \mathbb{E}[Y | \texttt{do}(X=x_c)]
\end{align}
where \texttt{do} denotes the intervention operation. In words, ATE aims to determine the relative expected difference in the value of $Y$ when we intervene $X$ to be $x_t$ compared to when we intervene $X$ to be $x_c$.  Similarly, consider the scenario in which we want to estimate the same ATE as above, but conditioned on some set of variables $C$ taking value $c$. Then CATE is defined as,
\begin{align}
    \label{eq_cate}
    \texttt{CATE} = \mathbb{E}[Y | \texttt{do}(X=x_t), C=c] - \mathbb{E}[Y | \texttt{do}(X=x_c), C=c]
\end{align}
Notice here that $X$ is intervened but $C$ is not.  Finally, consider the scenario where we want to estimate the outcome of an intervention on a specific sample (or unit). Specifically, suppose we have a specific instance of a system of random variables $(X_1, X_2,...,X_N)$ given by $(X_1=x_1, X_2=x_2,...,X_N=x_N)$, then in a counterfactual, we want to know the effect an intervention (say) $X_1=k$ would have had on some other variable(s) (say $X_2$), holding all the remaining variables fixed. Mathematically, this can be expressed as,

$$\texttt{Counterfactual} = X_2 | \texttt{do}(X_1=k), X_3=x_3, X_4=4,\cdots,X_N=x_N$$

Finding such effects from observational data alone can be challenging. Performing causal inference requires knowledge of the causal graph. There are existing methods for causal inference, for instance, using backdoor adjustment \citep{pearl1995causal} that involves finding a set of variables that result in non-causal associations between the treatment and target variable. In our current implementation, we support the backdoor method, and an in-house method in which we learn a set of relevant conditional models that are together able to simulate the data generating process, and we then use this process to estimate ATE, CATE and counterfactuals by performing interventions explicitly in this process.
\\\\
\textbf{Tabular}

Here we describe how our in-house method for causal inference works.
For tabular data, we first find the topologically sorted ordering of the variables in the causal graph. We then extract the causal paths from the treatment variables $X$ to the target variable $Y$. A causal path consists of a sequence of variables with directed edges. For each variable in these paths, we learn a statistical model (E.g. linear regression) that learns to predict this variable from all its parents from the given observational data. Once this is done, we once again traverse the causal graph in the topologically sorted order, and predict the value of each variable using the learned statistical models, except intervene the treatment variables $X$ whenever it is encountered. This process results in the estimated values of the target variable $Y$ under interventions. To compute ATE, we simply compute the mean under each intervention, and take the difference as shown in Eq. \ref{eq_ate}.

To compute CATE, in addition to the above process, we learn another statistical model that learns to predict the interventional value of the target variable $Y$ for each value of the condition variable $C$ in the observational data. We then simply use the desired value of the condition variable as input, to predict CATE. For couterfactuals, the procedure is similar to CATE, but instead of one or a few condition variables, we condition on all the variables except the target variable.
\\\\
\textbf{Time Series}

The causal inference process is performed similarly for the time series case, with one difference-- the data generation process is done iteratively as opposed to the tabular case, where all observations are generated at once in the topologically sorted order.

\section{Application: Root Cause Analysis (RCA)}

The RCA module supports two algorithms-- Root Cause Detector for time series data, and Tabular Distribution Shift Detector for tabular data.

\subsection{Root Cause Detector}

This algorithm detects root cause of anomaly in continuous time series data with the help of a higher-level context variable. The algorithm uses the PC algorithm to estimate the causal graph,
for root  cause analysis by treating the failure, represented using the higher-level metrics, as an intervention 
on the root cause node, and PC can use conditional independence tests to quickly detect 
which lower-level metric the failure node points to, as the root cause of anomaly \citep{ikram2022root,huang2020causal}.
This algorithm makes the following assumptions:  
(1) observational samples conditioned on the higher-level context variable (e.g., time index) are i.i.d., 
(2) linear relationship between variables with Gaussian noise terms, (3) Causal Markov condition, which implies that two variables that are d-separated in a causal graph are 
probabilistically independent, (4) faithfulness, i.e., no conditional independence can hold unless the Causal Markov condition is met, (5) no hidden confounders. 

\subsection{Tabular Distribution Shift Detector}

This algorithm detects the origins of distribution shifts in tabular, continuous/discrete data 
with the help of domain index variable. The algorithm uses the PC algorithm to estimate the causal graph,
by treating distribution shifts as intervention of the domain index on the root cause node, and PC can use 
conditional independence tests to quickly recover the causal graph and detect the root cause of anomaly \citep{ikram2022root}.
Note that the algorithm supports both discrete and continuous variables, and can handle nonlinear relationships
by converting the continuous variables into discrete ones using K-means clustering and using discrete PC algorithm
instead for CI test and causal discovery.
This algorithm makes the following assumptions: 
(1) observational samples conditioned on the domain index are i.i.d., 
(2) arbitrary relationship between variables, 
(3) Causal Markov condition, which implies that two variables that are d-separated in a causal graph are 
probabilistically independent, 
(4) faithfulness, i.e., no conditional independence can hold unless the Causal Markov condition is met, 
(5) no hidden confounders.

 \begin{figure}
    \centering
        \includegraphics[width=0.45\columnwidth]{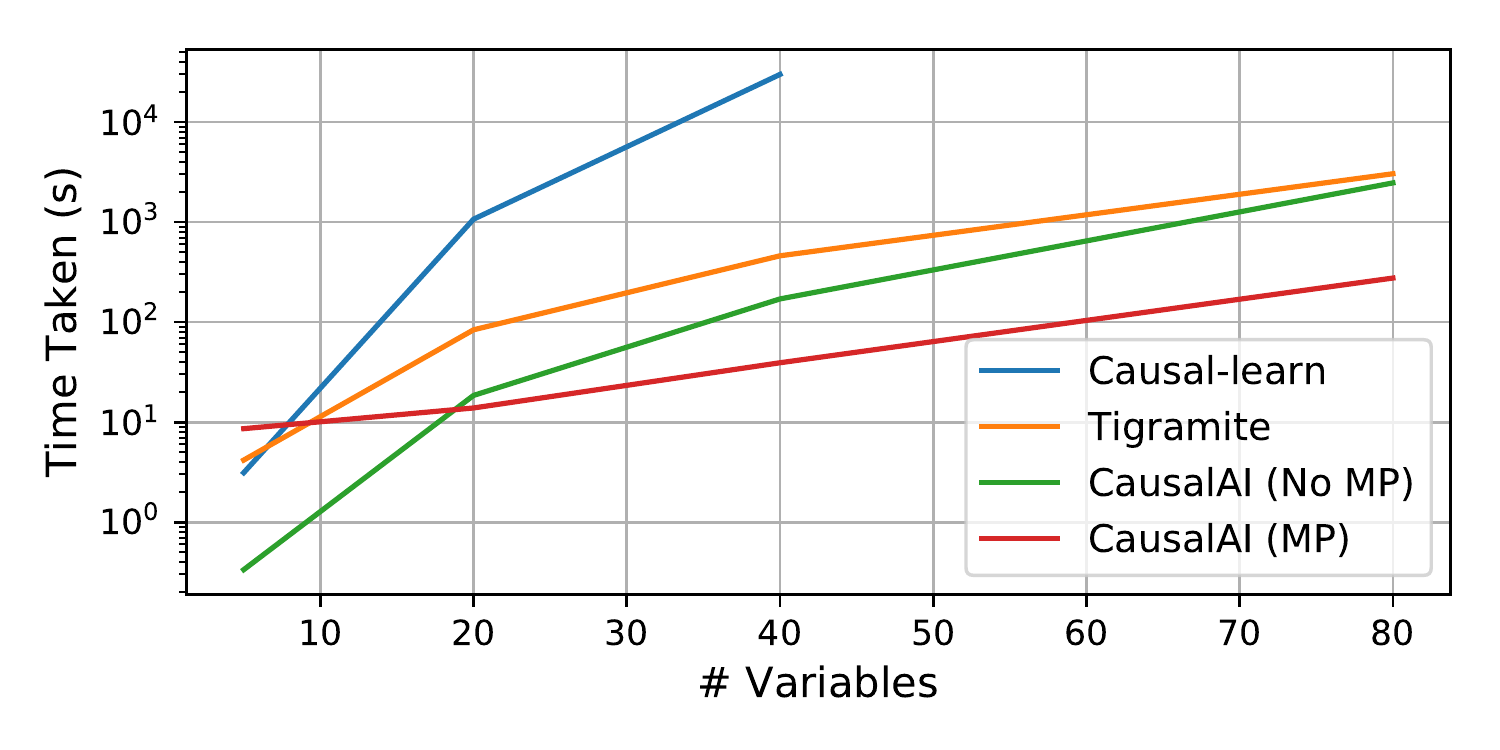}
        \includegraphics[width=0.45\columnwidth]{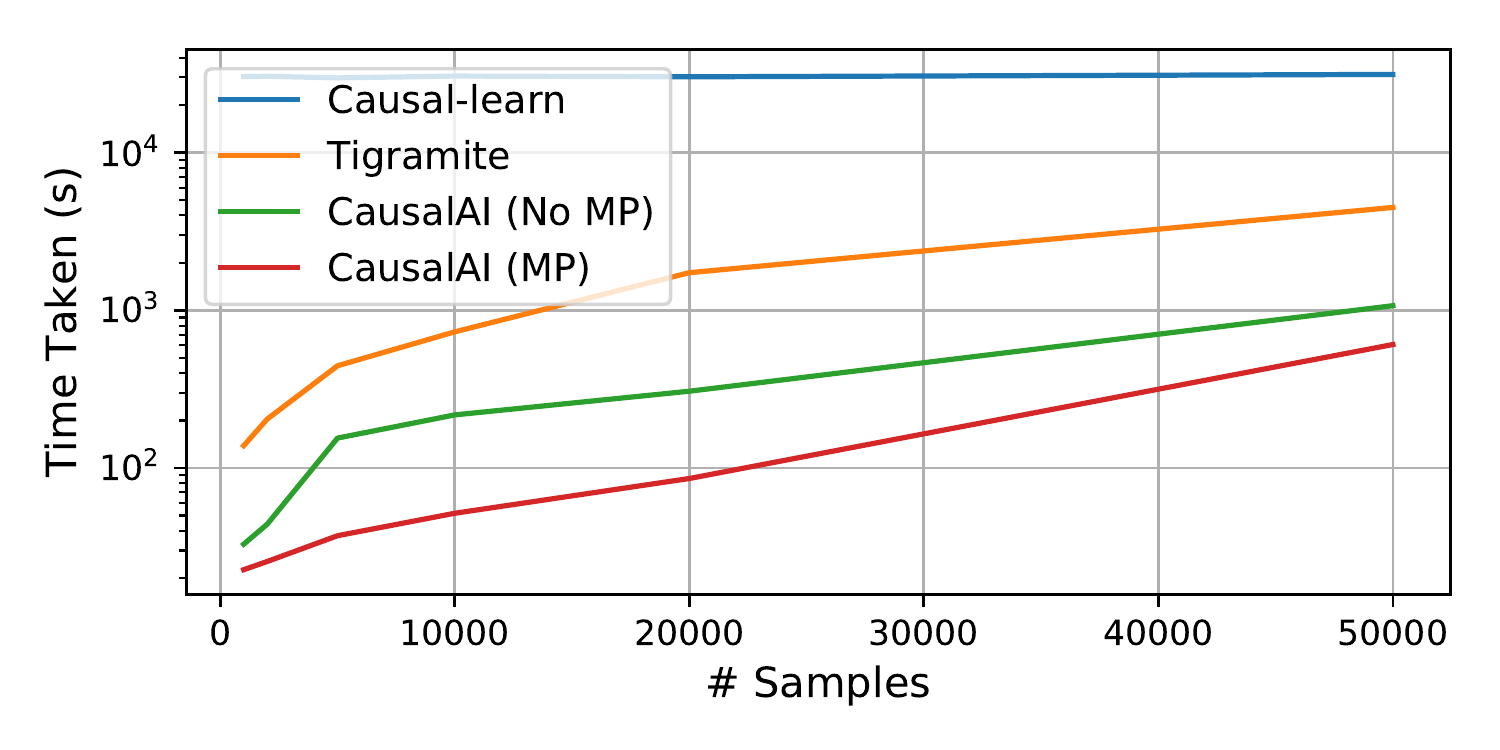}
    \caption{A comparison of speed (seconds) of the PC algorithm implementation of our library with existing libraries with varying number of variables (left) and samples (right). CausalAI (MP) indicates the PC method uses multi-processing, while CausalAI(No MP) indicates no multi-processing.}
    \label{fig:speed}
\end{figure}

 \begin{figure}
    \centering
        \includegraphics[width=0.45\columnwidth]{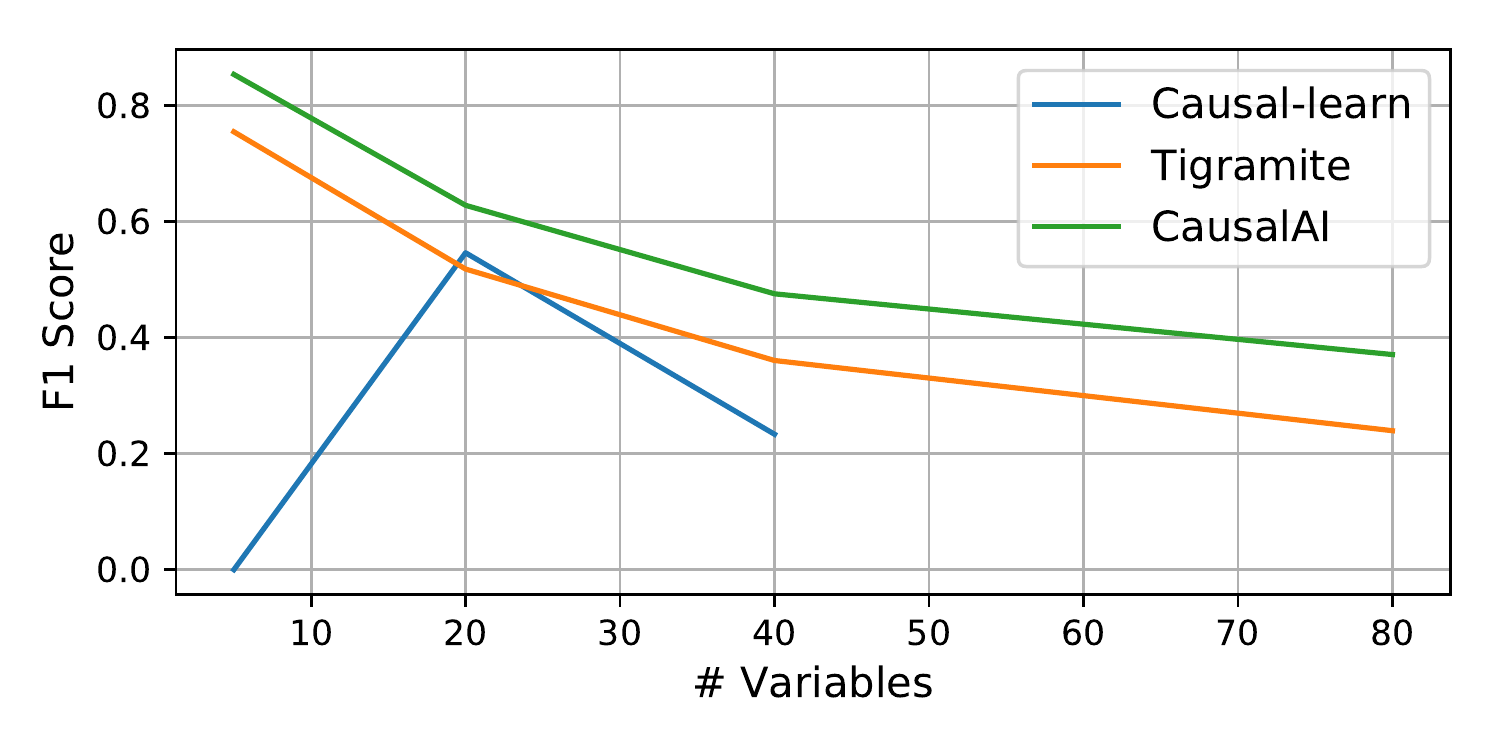}
        \includegraphics[width=0.45\columnwidth]{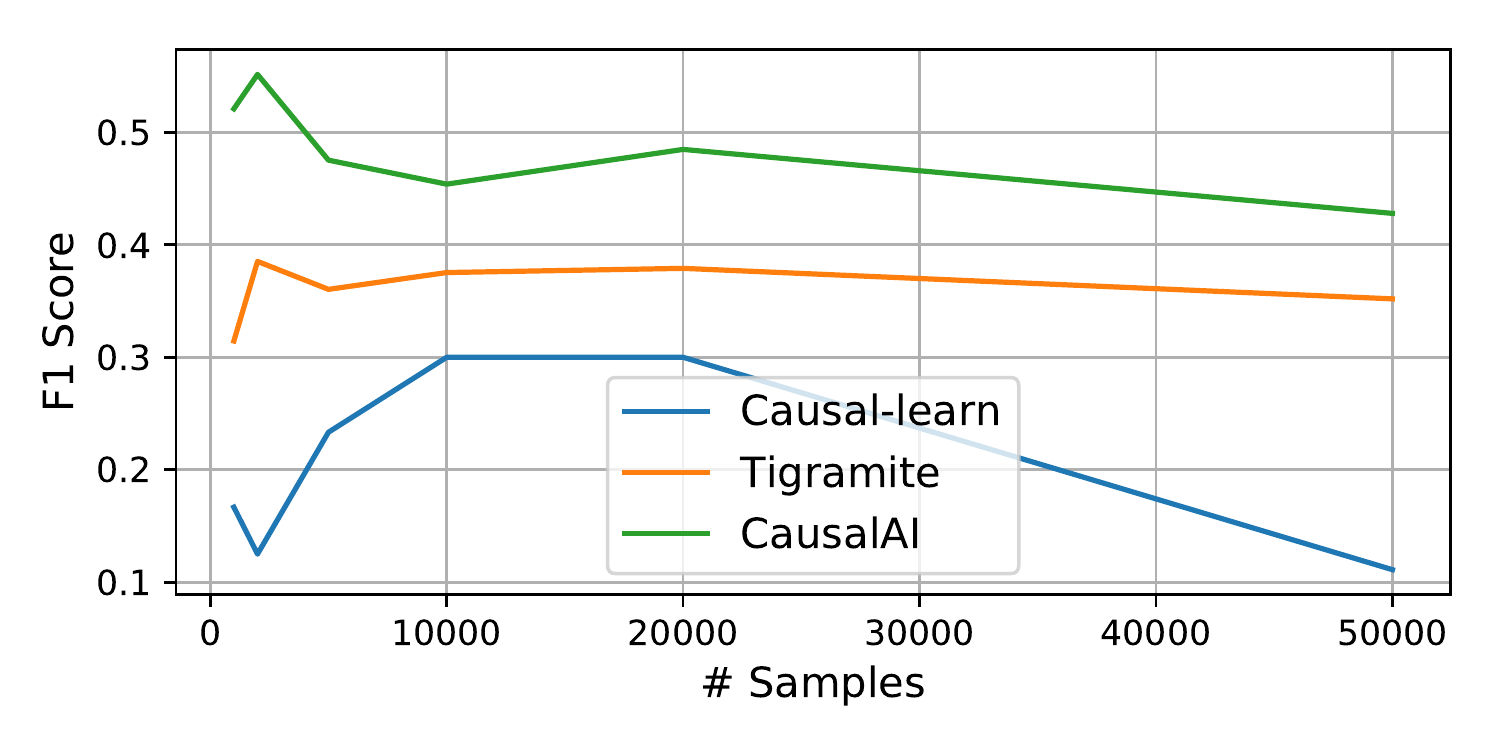}
    \caption{A comparison of the F1 score achieved by the PC algorithm implementation of our library with existing libraries with varying number of variables (left) and samples (right).}
    \label{fig:f1}
\end{figure}

\section{Experiments}

\subsection{Causal Discovery for Time Series}

\textbf{Experimental Settings} We conduct empirical studies on synthetic data to verify the efficiency of CausalAI library. Specifically, We compare PC algorithm in CausalAI library with hat in \textit{causal-learn}\footnote{https://github.com/cmu-phil/causal-learn} library and \textit{tigramite} library\footnote{https://github.com/jakobrunge/tigramite}. To achieve fair comparison, the p-value threshold and maximum condition set size are set to $0.05$ and $4$. We report results on two settings, considering the effect of different number of variables and the effect of different number of samples, respectively. We evaluate the training time and F1 score of the estimated graphs.

\noindent\textbf{Results} Figure \ref{fig:speed} reports the time cost required by each model. We can find that CausalAI and tigramite are much faster than causal-learn. This is because both of these two libraries consider the characteristics of time series and reduce the potential search space when uncovering the causal relationships. We also observe a speedup via using multi-processing in CausalAI, especially when handling data with large sample size or large variable size. This verifies the efficacy of the proposed method in CausalAI to handle large scale data. 
 
 In Figure \ref{fig:f1}, we also show the F1 score between the estimated causal graphs and the ground truth causal graph. We find that CausalAI and tigramite performs much better than causal-learn. This is because the PC algorithm in causal-learn is a more general implementation without considering the characteristic of time series. Therefore, it is necessary to manually include suitable prior knowledge to further improve its performance.

\section{Conclusions and Future Work}

We introduce the Salesforce CausalAI Library, an open source library for causal analysis of time series and tabular data. The Salesforce CausalAI Library aims to provide a one-stop solution to the various needs in causal analysis including handling different data types, data generation, multi-processing for speed-up, utilizing domain knowledge and providing a user-friendly code-free visual interface.

We continue to develop this library and invite the research community to contribute by submitting pull requests to our GitHub repository. Some of our future plans are to include support for deep learning based causal discovery algorithms and more applications of causal discovery algorithms in addition to root cause analysis.









\section{Contributions}

\textbf{Devansh Arpit}: Implemented the code for Salesforce CausalAI Library v1.0 and v2.0, except modules covered by other authors mentioned below. Created the Sphinx documentation for the GitHub repository. Wrote the blog. Wrote this tech report, except for the experiments section.\\\\
\textbf{Matthew Fernandez}: Implemented the UI to test and showcase the CausalAI Library. \\\\
\textbf{Itai Feigenbaum}: Coded Markov Blanket Discovery algorithm Grow-Shrink and helped improve the broader codebase.\\\\
\textbf{Weiran Yao}: Coded the root cause analysis and distribution shift detection algorithms.\\\\
\textbf{Chenghao Liu}: Implemented the VARLINGAM algorithm, wrote the experiments section and helped in the literature review.\\\\
\textbf{Wenzhuo Yang}: Coded algorithm which will be added to the next version of the CausalAI library.\\\\
\textbf{Paul Josel}: Helped in the UI design and organization process.\\\\
\textbf{Shelby Heinecke}: Contributed to the conception and initial direction of the library. Contributed to the literature review.\\\\
\textbf{Eric Hu}: Coordinated UI design and implementation.\\\\
\textbf{Huan Wang}: Contributed to the library design regarding tabular/time series data structures, choices of algorithms, and UI features. \\\\
\textbf{Stephen Hoi}: Contributed to the high-level discussions on the project. General feedback for the project and the report.\\\\
\textbf{Caiming Xiong}: Initiated project idea and contributed to the high-level direction of the library. General project feedback.\\\\
\textbf{Kun Zhang}: Contributed to causal discovery algorithm selection and design. General project feedback.\\\\
\textbf{Juan Carlos Niebles}: Contributed to the high-level project direction and scope, oversaw project execution and coordinated team effort.\\\\

\vskip 0.2in
\clearpage
\bibliography{ref}

\end{document}